# A Translational 3DOF Parallel Mechanism with Partial Motion Decoupling and Analytic Direct Kinematics[1]

**Huiping Shen**
School of Mechanical Engineering,
Changzhou University,
Changzhou 213016, China
shp65@126.com

**Damien Chablat**
CNRS, Laboratoire des Sciences du Numérique de Nantes,
UMR 6004 Nantes, France
Damien.Chablat@cnrs.fr

**Boxiong Zeng**
School of Mechanical Engineering,
Changzhou University,
Changzhou 213016, China
710022180@qq.com

**Ju Li**
Changzhou University, School of Mechanical Engineering,
Changzhou 213164, China.
wangju0209@163.com

**Guanglei Wu**
Dalian University of Technology, School of Mechanical Engineering,
Dalian 116024, China
gwu@dlut.edu.cn

**Ting-Li Yang**
School of Mechanical Engineering,
Changzhou University,
Changzhou 213016, China
yangtl@126.com

## ABSTRACT

According to the topological design theory and method of parallel mechanism (PM) based on position and orientation characteristic (POC) equations, this paper studied a 3−DOF translational PM that has three advantages, i.e., (i) it consists of three fixed actuated prismatic joints, (ii) the PM has analytic solutions to the direct and inverse kinematic problems, and (iii) the PM is of partial motion decoupling property. Firstly, the main topological characteristics, such as the POC, degree of freedom and coupling degree were calculated for kinematic modeling. Thanks to these properties, the direct and inverse kinematic problems can be readily solved. Further, the conditions of the singular configurations of the PM were analyzed which corresponds to its partial motion decoupling property.

## INTRODUCTION

In many industrial production lines, process operations require only pure translational motions. Therefore, the 3-DOF translational parallel mechanism (TPM) has a significant potential, thanks to its fewer actuated joints, a relatively simple structure and easy to be controlled. Many researchers have studied the TPMs. For example, original design of 3-DOF TPM is the Delta Robot, which was presented by Clavel [1]. The Delta-based TPMs have been developed with prismatic actuated joints [2,

---

3] and the sub-chain is a 4R parallelogram mechanism[2]. Several optimizations based on the Jacobian matrices were carried out in [4-7]. In [8, 9], the authors proposed a 3-RRC TPM and developed the kinematics and workspace analysis. Kong *et al.* [10] proposed a 3-CRR mechanism with good motion performance and no singular postures. Li *et al.*, [11, 12] developed a 3-UPU PM and analyzed the instantaneous kinematic performance of the TPM. Yu *et al.* [13] carried out a comprehensive analysis of the three-dimensional TPM configuration based on the screw theory. Lu *et al.* [14] proposed a 3-RRRP (4R) three-translation PM and analyzed the kinematics and workspace. Yang *et al.* [15, 16] studied 3T0R PMs based on the single opened chains (SOC) units, and a variety of new TPMs were synthesized and then classified [17, 18]. Considering the anisotropy of kinematics, Zhao *et al.* [19] analyzed the dimensional synthesis and kinematics of the 3-DOF translational Delta PM. Zeng *et al.* [20-22] introduced a 3-DOF TPM called as Tri-pyramid robot and presented a more detailed analytical approach for the Jacobian matrix. Prause *et al.* [23] compared the characteristics of dimensional synthesis, boundary conditions and workspace for various 3-DOF TPMs for the best performances among them. Mazare *et al.* [24] proposed a 3-DOF 3-[P2(US)] mechanism and analyzed its kinematics and dexterity.

However, most of the previous TPMs generally suffer from two major problems: i) the coupling degree κ of these PMs is greater than zero, which means that analytical direct position solution is difficult to be derived, and ii) these PMs do not have input-output decoupling characteristics [25], leading to the complexity of motion control and path planning.

In the next sections, the topology design theory of PM based on position and orientation characteristics (POC) equations [16, 17] is applied to present a new TPM, with the coupling degree determined. Then, the position analysis is conducted to deduce the direct and inverse kinematic model. For a given example, the maximum number of solutions for the direct and inverse kinematic model are obtained. Then, the configurations for the serial and parallel singularities are presented.

## DESIGN AND TOPOLOGY ANALYSIS

### Topological design

The 3T parallel manipulator proposed in this paper is illustrated in Fig. 1. The base platform 0 is connected to the moving platform 1, by two hybrid chains that contain both loop(s) and serial joints. A chain that is composed of links and joints in series is called Single-Opened-Chains (SOCs), while hybrid chains are called hybrid Single-Opened-Chains (HSOCs). The structural and geometric constraints of two HSOCs are given as follows:

---

[2] Throughout this paper, R, C, P and U stand for revolute, cylindrical, prismatic and universal joints, respectively.

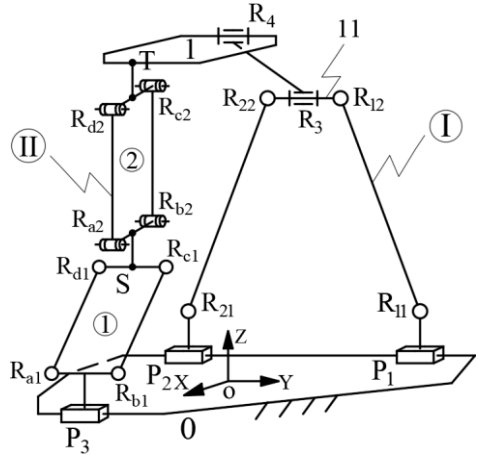

FIGURE 1 KINEMATIC STRUCTURE OF THE 3T PM

- For the 6-bar planar mechanism loop (abbreviation: 2P4R planar mechanism) in right side of Fig. 1, two revolute joints $R_3$ and $R_4$ with axes parallel to each other are connected in series, where $R_3$ is connected to link 11 and $R_4$ is connected to the moving platform 1 to obtain the first HSOC branch (denoted as: hybrid chain I). Two prismatic joints $P_1$ and $P_2$ of the 6-bar planar mechanism will be used as actuated joints.

- The left side branch is made up of a prismatic joint $P_3$ and two 4R parallelogram mechanisms connected in series, and the parallelograms connected from $P_3$ to the moving platform 1 are respectively recorded as ①, ②. The prismatic joint, $P_3$ and the parallelogram ① are rigidly connected with the motion confined in the same plane, and they are connected to the parallelogram ② in their orthogonal plane to obtain the second HSOC branch (denoted as: hybrid chain II).

- The prismatic joints $P_1$, $P_2$ and $P_3$ are connected to the base platform 0; $P_1$ and $P_2$ are arranged coaxially, and prismatic $P_1$ is parallel to $P_3$. When the PM moves, the 2P4R planar mechanism is always parallel to the plane of the parallelogram ①.

**Analysis of topology characteristics**

*Analysis of the POC set*: The POC set equations for parallel mechanisms are expressed, respectively, as follows [16]:

$$M_{bi} = \bigcup_{i=1}^{m} M_{Ji} \tag{1}$$

$$M_{Pa} = \bigcap_{i=1}^{n} M_{bi} \tag{2}$$

where

- $M_{Ji}$ - POC set generated by the $i$-th joint.

- $M_{bi}$ - POC set generated by the end link of $i$-th branched chain.

- $M_{Pa}$ - POC set generated by the moving platform of PM.

- POC- position and orientation characteristics
- ∪-union operation
- ∩-intersection operation

Apparently, the output motions of the intermediate link 11 in the 2P4R planar mechanism on the hybrid chain I are two translations and one rotation (2T1R), hence, the output motions of the end link of the hybrid chain I are three translations and two rotations (3T2R).

The output motions of the link S of the parallelogram ① on the hybrid chain II are two translations (2T), thus, the output motions of the link T of the parallelogram ② on the hybrid chain II are three translations (3T).

Therefore, the topological architecture of the hybrid chain I and II of the PM can be equivalently denoted as [16], respectively:

$$HSOC_1\left\{-(P^{(2P4R)}-P^{(2P4R)}) \perp R^{(2P4R)} \perp R_3 \parallel R_4-\right\}$$

$$HSOC_2\left\{-P_3 - P^{(4R)} - P^{(4R)} -\right\}$$

Where,

- $P^{(2P4R)}-P^{(2P4R)} \perp R^{(2P4R)}$ means that the 2P4R planar mechanism generates two translations and one rotation, which is denoted as $\begin{bmatrix} t^2(\perp R_{12}) \\ r^1(\parallel R_{12}) \end{bmatrix}$, while $\left\{-P^{(4R)} - P^{(4R)} -\right\}$ means two translations generated by two parallelograms composed of 4R joints that is denoted as $\begin{bmatrix} t^2 \\ r^0 \end{bmatrix}$.

- $t^2(\perp R_{12})$ means that there are two translations, i.e., $t^2$, in the plane that is perpendicular to the axis of joint $R_{12}$. Moreover, $r^1(\parallel R_{12})$ means that there is one rotation, i.e., $r^1$, that is parallel to the axis of joint $R_{12}$. The other notations in the formulas above can be found in [16].

The POC sets of the end link of the two HSOCs are determined according to Eq. (1) as follows:

$$M_{HSOC_1} = \begin{bmatrix} t^2(\perp R_{12}) \\ r^1(\parallel R_{12}) \end{bmatrix} \cup \begin{bmatrix} t^2(\perp R_3) \\ r^1(\parallel R_3) \end{bmatrix} = \begin{bmatrix} t^3 \\ r^2(\parallel \Diamond(R_{12}, R_3)) \end{bmatrix}$$

$$M_{HSOC_2} = \begin{bmatrix} t^1(\parallel P_3) \\ r^0 \end{bmatrix} \cup \begin{bmatrix} t^2 \\ r^0 \end{bmatrix} = \begin{bmatrix} t^3 \\ r^0 \end{bmatrix}$$

The POC set of the moving platform of this PM is determined from Eq. (2) by

$$M_{Pa} = M_{HSOC_1} \cap M_{HSOC_2} = \begin{bmatrix} t^3 \\ r^0 \end{bmatrix}$$

This formula indicates that the moving platform 1 of the PM produces three-translation motion. It is further known that the hybrid chain II in the mechanism itself can realize the design requirement of three translations, which simultaneously constrains the two rotational outputs of the hybrid chain I.

*Determining the DOF:* The general and full-cycle DOF formula for PMs proposed in author's work [16] is given below:

$$F = \sum_{i=1}^{m} f_i - \sum_{j=1}^{v} \xi_{Lj} \tag{3}$$

$$\sum_{j=1}^{v} \xi_{Lj} = \dim\left\{ (\bigcap_{i=1}^{j} M_{b_i}) \mathrm{Y} M_{b_{(j+1)}} \right\} \tag{4}$$

where

- $F$ - DOF of PM.
- $f_i$ - DOF of the $i^{\text{th}}$ joint.
- $m$ - number of all joints of the PM.
- $v$ - number of independent loops of the PM, and $v = m - n + 1$.
- $n$ - number of links.
- $\xi_{L_j}$ - number of independent equations of the $j^{\text{th}}$ loop.
- $\bigcap_{i=1}^{j} M_{b_i}$ - POC set generated by the sub-PM formed by the former $j$ branches.
- $M_{b_{(j+1)}}$ - POC set generated by the end link of $j+1$ sub-chains.

The PM can be decomposed into two independent loops, and their constraint equations are calculated as follows:

① The first independent loop is consisted of the 2P4R planar mechanism in the hybrid chain I, the $LOOP_1$ is deduced as:

$$LOOP_1\left\{-(P^{(2P4R)}, P^{(2P4R)}) \perp R^{(2P4R)} -\right\}$$

Obviously, the independent displacement equation number of the planar mechanism is $\xi_{L_1} = 3$.

② The above 2P4R planar mechanism and the following sub-string $R_3 \| R_4$ with the additional $HSOC_2$ will form the second independent loop, namely,

$$LOOP_2\left\{-R_3 \| R_4 - P^{(4R)} - P^{(4R)} - P_3 -\right\}$$

In accordance with Eq.(4), the independent displacement equation number $\xi_{L_2}$ of the second loop can be obtained [16] as below:

$$\xi_{L_2} = \dim.\left\{ \begin{bmatrix} t^2(\perp R_{12}) \\ r^1(\| R_{12}) \end{bmatrix} \mathrm{U} \begin{bmatrix} t^3 \\ r^1(\| R_3) \end{bmatrix} \right\} = \dim.\left\{ \begin{bmatrix} t^3 \\ r^2(\| \Diamond(R_{12}, R_3)) \end{bmatrix} \right\} = 5$$

Thus, the DOF of the PM is calculated from Eq. (3) expressed as

$$F = \sum_{i=1}^{m} f_i - \sum_{j=1}^{2} \xi_{L_j} = (6+5) - (3+5) = 3$$

Therefore, the DOF of the PM is equal to 3, and when the prismatic joints $P_1$, $P_2$ and $P_3$ on the base platform 0 are the actuated joints, the moving platform 1 can realize three-translational motion.

*Determining the coupling degree:* According to the composition principle of mechanism based on single-opened-chains (SOC) units, any PM can be decomposed into a series of Assur kinematic chains (AKC), and an AKC with $v$ independent loops can be decomposed into $v$ SOC. The constraint degree of the $j^{th}$ SOC, $\Delta_j$, is defined [16, 17] by

$$\Delta_j = \sum_{i=1}^{m_j} f_i - I_j - \xi_{L_j} = \begin{cases} \Delta_j^- = -5, -4, -2, -1 \\ \Delta_j^0 = 0 \\ \Delta_j^+ = +1, +2, +3, \cdots \end{cases} \quad (5)$$

where

- $\Delta_j$ - constraint degree of the $j^{th}$ SOC.
- $m_j$ - number of joints contained in the $j^{th}$ SOC$_j$.
- $f_i$ - DOF of the $i^{th}$ joints.
- $I_j$ - number of actuated joints in the $j^{th}$ SOC$_j$.
- $\xi_{L_j}$ - number of independent equations of the $j^{th}$ loop.

For an AKC, it must be satisfied with the following equation.

$$\sum_{j=1}^{v} \Delta_j = 0$$

Sequentially, the coupling degree of AKC [16, 17] is defined by

$$\kappa = \frac{1}{2} \min \left\{ \sum_{j=1}^{v} |\Delta_j| \right\} \quad (6)$$

The physical meaning of the coupling degree $\kappa$ can be intepreted in this way. The coupling degree $k$ describes the complexity level of the topological structure of a PM, and it also represents the complexity level of its kinematic and dynamic analysis. It has been proved that the higher the coupling degree $\kappa$ is, the more complex the kinematic and dynamic solutions of the PM are [15, 16].

The number of independent displacement equations of $LOOP_1$ and $LOOP_2$ have been calculated in the previous section *Determining the DOF*, i.e., $\xi_{L_1} = 3$, $\xi_{L_2} = 5$, thus, the constraint degree of the two independent loops are calculated by Eq. (5), respectively, with the solution below:

$$\Delta_1 = \sum_{i=1}^{m_1} f_i - I_1 - \xi_{L_1} = 6 - 2 - 3 = 1, \quad \Delta_2 = \sum_{i=1}^{m_2} f_i - I_2 - \xi_{L_2} = 5 - 1 - 5 = -1$$

The coupling degrees of the AKC is calculated by Eq. (6) as

$$k = \frac{1}{2} \sum_{j=1}^{v} |\Delta_j| = \frac{1}{2} (|+1| + |-1|) = 1$$

Thus, the PM contains only one AKC, and its coupling degrees is equal to 1. Therefore, when solving the direct position solutions of the PM, it is necessary to assign only one virtual variable in the $1^{st}$ loop whose constraint degree is one ($\Delta_j = +1$). Then, one constraint equation with this virtual variable is established in the second loop with the negative constraint degree ($\Delta_j = -1$). Further, the real value of this virtual variable can be obtained by the numerical method with only one unknown, thus, the direct position solutions of the PM are obtained finally.

However, owing to the 3-translational output motion, a very special constraint, of the moving platform 1 of the PM, the second loop with the negative constraint degree ($\Delta_j = -1$) can be directly applied to the geometric constraint of the first loop with the positive constraint degree is one ($\Delta_j = +1$). That means that the motion of the link 11 is always parallel to the base platform 0, which can be easily determined from the second loop. Therefore, the virtual variable is easily obtained from the first loop, and there is no need to solve the virtual variable by one-dimensional numerical method, which significantly simplifies the process of the direct solutions. Thus, the analytical direct position solutions of the PM can be directly obtained in the following section.

## POSITION ANALYSIS

### The coordinate system and parameterization

The kinematic modeling of the PM is shown in Fig. 2. The base platform 0 is in a rectangular shape with a length and a width of $2a$ and $2b$, respectively. The global coordinate system O-XYZ is established on the base platform 0, with the origin at the geometric center. The X and Y axes are perpendicular and parallel to the line $A_1A_2$, and the Z axis is normal to the base plane pointing upwards. The moving coordinate system O'-X'Y'Z' is established with the coordinate axes parallel to those of the global frame located at the geometric center of the moving platform.

The length of the three driving links 2 is equal to $l_1$, the lengths of the connecting links 9 and 10 on the hybrid chain I are both equal to $l_2$, and the lengths of the intermediate links 11 and 12 are equal to $l_3$, $l_4$, respectively.

The length of the parallelogram short links 3, 6 on the hybrid chain II is equal to $l_5$. The point $B_3$, $C_3$, $D_3$ and $E_3$ are the midpoints of the short edges, respectively. The length of the long links 4, 7 is equal to $l_6$, and the length of the connecting link 5 between the parallelograms is $l_7$. Moreover, the length of the connecting link 8 is $l_8$, and the length of the line $D_2F_3$ on the moving platform 1 is $2d$.

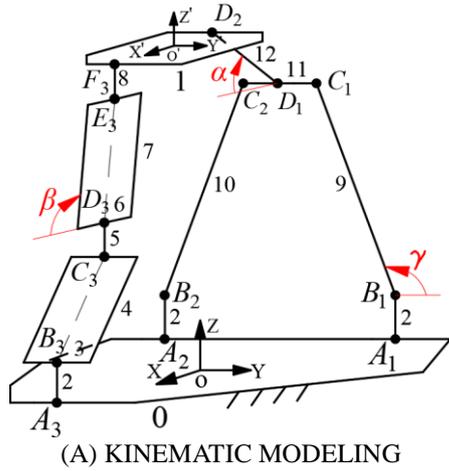
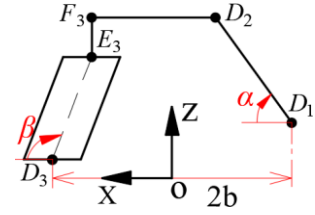

(A) KINEMATIC MODELING  (B) GEOMETRIC RELATIONSHIP OF THE SECOND LOOP (PARTIAL) IN THE XOZ DIRECTION

FIGURE 2 KINEMATIC MODELING OF THE 3T PM

The angle between the vectors $B_1C_1$ and the Y axis is $\gamma$, and the $\gamma$ is assigned as virtual variable. The angles between the vectors $D_1D_2$, $D_3E_3$ and the X axis are $\alpha$ and $\beta$, respectively.

**Direct kinematic problem**

To solve the direct kinematic problem, it is to compute the position $O'(x,y,z)$ of the moving platform when setting the position coordinates of the prismatic joints at points $P_1$, $P_2$ and $P_3$, with the coordinates $y_{A_1}$, $y_{A_2}$ and $y_{A_3}$.

1) **Solving the first loop**

$$LOOP_1: A_1 - B_1 - C_1 - C_2 - B_2 - A_2$$

The coordinates of points $A_1$, $A_2$ and $A_3$ on the base platform 0 are derived, respectively

$$A_1 = (-b, y_{A_1}, 0)^T, A_2 = (-b, y_{A_2}, 0)^T, A_3 = (b, y_{A_3}, 0)^T.$$

The coordinates of the three revolute joints, $B_1$, $B_2$ and $B_3$ will be solved as

$$B_1 = (-b, y_{A_1}, l_1)^T, B_2 = (-b, y_{A_2}, l_1)^T, B_3 = (b, y_{A_3}, l_1)^T.$$

Due to the special constraint of the three translations of the moving platform 1, during the movement of the PM, the intermediate link 11 of the 2P4R planar mechanism is always parallel to the base platform 0, that is, $(C_1C_2) \| (A_1A_2)$. Then the following constraint equation produced by topological characteristics of three-translation outputs of the moving platform is obtained.

$$z_{C_1} = z_{C_2} \qquad (7)$$

Therefore, the coordinates of points $C_1$ and $C_2$ are calculated as

$$C_1 = (-b, y_{A_1} + l_2 \cos\gamma, l_1 + l_2 \sin\gamma)^T \text{ and } C_2 = (-b, y_{A_1} - l_2 \cos\gamma - l_3, l_1 + l_2 \sin\gamma)^T$$

With the link length constraints defined by $B_2C_2 = l_2$, two constraint equations can be deduced as below,

$$(x_{C_1} - x_{B_1})^2 + (y_{C_1} - y_{B_1})^2 + (z_{C_1} - z_{B_1})^2 = l_2^2$$

$$(x_{C_2} - x_{B_2})^2 + (y_{C_2} - y_{B_2})^2 + (z_{C_2} - z_{B_2})^2 = l_2^2 \qquad (8)$$

Equation (8) leads to $AB\cos\gamma + B^2 = 0$. The value of $\gamma$ can be determined as long as $A \neq 0$, namely,

$$\gamma = \pi \pm \arccos\left(\frac{B}{A}\right) \text{ with } A = 2l_2, B = y_{A_1} + l_3 - y_{A_2}. \qquad (9)$$

From the first loop, there are two solutions for this PM due to the two existng intersection points of two circles defined by Eq. (8). Thus, the second loop acts on the special geometric constraint of the Eq. (7) on the first loop, which is the key to directly finding the analytical solutions of $\gamma$. This is an advantage to ease the derivation of the analytical direct kinematics of this PM.

2) **Solving the second loop**

$$LOOP_2 : D_1 - D_2 - F_3 - E_3 - D_3 - C_3 - B_3 - A_3$$

The coordinates of points $D_1$ and $D_2$ obtained from points $C_1$ and $C_2$ are calculated as

$$D_1 = \begin{bmatrix} -b \\ y_{A_1} + l_2 \cos\gamma - l_3/2 \\ l_1 + l_2 \sin\gamma \end{bmatrix} \text{ and } D_2 = \begin{bmatrix} -b + l_4 \cos\alpha \\ y_{A_1} + l_2 \cos\gamma - l_3/2 \\ l_1 + l_2 \sin\gamma + l_4 \sin\alpha \end{bmatrix}$$

Simultaneously, the coordinates of point $O'$ can be calculated as:

$$O' = \begin{bmatrix} x \\ y \\ z \end{bmatrix} = \begin{bmatrix} -b + l_4 \cos\alpha + d \\ y_{A_1} + l_2 \cos\gamma - l_3/2 \\ l_1 + l_2 \sin\gamma + l_4 \sin\alpha \end{bmatrix} \qquad (10)$$

Further, the coordinates of points $F_3$, $E_3$, $D_3$ and $C_3$ are represented with the known coordinates of point $O'$ as below:

$$F_3 = (x+d, y, z)^T$$
$$E_3 = (x+d, y, z-l_8)^T$$
$$D_3 = (b, y, z-l_8 - l_6 \sin\beta)^T$$
$$C_3 = (b, y, z-l_8 - l_6 \sin\beta - l_7)^T \qquad (11)$$

With the link length constraints defined by $B_3C_3 = l_6$, the constraint equation can be deduced as below.

$$(x_{C_3} - x_{B_3})^2 + (y_{C_3} - y_{B_3})^2 + (z_{C_3} - z_{B_3})^2 = l_6^2 \qquad (12)$$

from Fig. 2, with the following equation, ,

$$l_4 \sin\alpha - l_6 \sin\beta = t \qquad (13)$$

there exists

$$(H_1 + t)^2 - H_2 = 0$$
$$t = -H_1 \pm \sqrt{H_2} \qquad (14)$$

with

$$H_1 = l_2 \sin\gamma - l_8 - l_7, \quad H_2 = l_6^2 - (y_{C_3} - y_{B_3})^2.$$

When the PM moves, the 2P4R planar mechanism is always parallel with the plane of the parallelogram ①, therefore, the following relationship always exists:

$$y_{D_1} = y_{D_3} \tag{15}$$

$$l_4 \cos\alpha + 2d - l_6 \cos\beta = 2b \tag{16}$$

Eliminating $\beta$ from Eqs. (13) and (16), yielding

$$J_1 \sin\alpha + J_2 \cos\alpha + J_3 = 0$$

$$\alpha = \arctan\left(\frac{J_1 J_3 + J_2 \sqrt{J_1^2 + J_2^2 - J_3^2}}{J_2 J_3 - J_1 \sqrt{J_1^2 + J_2^2 - J_3^2}}\right) \text{ and } \alpha = \arctan\left(\frac{J_1 J_3 - J_2 \sqrt{J_1^2 + J_2^2 - J_3^2}}{J_2 J_3 + J_1 \sqrt{J_1^2 + J_2^2 - J_3^2}}\right) \tag{17}$$

Where $J_1 = 2l_4 t$, $J_2 = 4l_4(b-d)$, $J_3 = l_6^2 - l_4^2 - t^2 - 4(b-d)^2$. To this end, by substituting the values of $\gamma$ and $\alpha$ obtained from Eqs. (9) and (17) into Eq. (10), the coordinates of point $O'$ in the reference coordinate system can be obtained, namely,

$$\begin{cases} x = f_1'(y_{A_1}, y_{A_2}, y_{A_3}) \\ y = f_2'(y_{A_1}, y_{A_2}) \\ z = f_3'(y_{A_1}, y_{A_2}, y_{A_3}) \end{cases}$$

Thus, the PM has partial input-output motion decoupling, which is advantageous for trajectory planning and motion control of the moving platform. For the sake of understanding, the above calculation procedure can be depicted in Fig. 3.

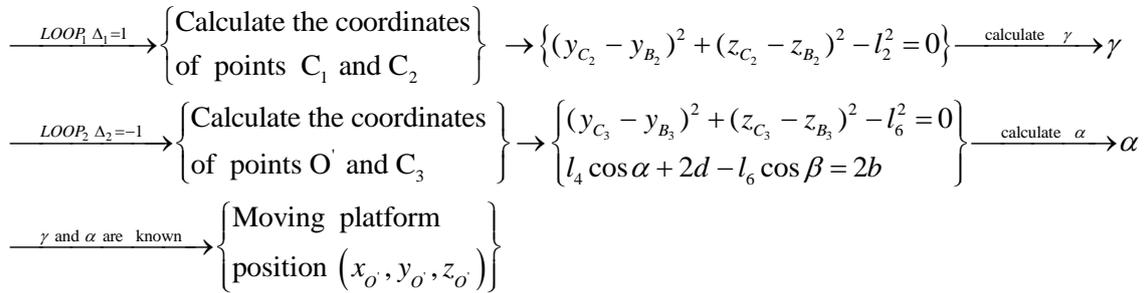

FIGURE 3 FLOW CHART OF DIRECT POSITION SOLUTIONS

It can be seen that the geometric constraints Eqs. (7), (15) and (16) are the key to find the analytical equation of the first and second loop position equations of the PM. In summary, when the positions of three prismatic joints $P_1, P_2, P_3$ are know, the mechanism under study can have up to eight direct kinematic solutions, according to Eqs. (9), (16) and (17).

**Inverse kinematic problem**

To solve the inverse kinematics, the values of $y_{A_1}$, $y_{A_2}$ and $y_{A_3}$ as a function of the coordinate $O'(x,y,z)$ of the moving platform are computed. For a given position of the moving platform, from Eqs. (10) and (16), the angles $\alpha$ and $\beta$ are calculated as

$$\alpha = \pm \arccos\left(\frac{x+b-d}{l_4}\right) \tag{18}$$

$$\beta = \pm \arccos\left(\frac{x+d-b}{l_6}\right) \tag{19}$$

Further, the coordinates of points $C_1$ and $C_2$ are defined as:

$$C_1 = (-b, y + l_3/2, z - l_4 \sin\alpha)^T$$
$$C_2 = (-b, y - l_3/2, z - l_4 \sin\alpha)^T$$

In addition, the coordinates of point $C_3$ have been given by Eq. (11). Therefore, with the link length constraints defined by $B_1C_1 = B_2C_2 = l_2$ and $B_3C_3 = l_6$, there are three constraint equations as below.

$$\begin{cases} (x_{C_1} - x_{B_1})^2 + (y_{C_1} - y_{B_1})^2 + (z_{C_1} - z_{B_1})^2 = l_2^2 \\ (x_{C_2} - x_{B_2})^2 + (y_{C_2} - y_{B_2})^2 + (z_{C_2} - z_{B_2})^2 = l_2^2 \\ (x_{C_3} - x_{B_3})^2 + (y_{C_3} - y_{B_3})^2 + (z_{C_3} - z_{B_3})^2 = l_6^2 \end{cases} \tag{20}$$

From Eqs. (20), $y_{A_i}$ ($i=1,2,3$) are calculated as following.

$$y_{A_i} = y_{C_i} \pm \sqrt{M_i} \quad (i=1,2,3) \tag{21}$$

with

$$M_1 = l_2^2 - (z_{C_1} - l_1)^2, \; M_2 = l_2^2 - (z_{C_2} - l_1)^2, \; M_3 = l_6^2 - (z_{C_3} - l_1)^2.$$

In summary, when the coordinates of point $O'$ on the moving platform 1 are known, each input values $y_{A_1}$, $y_{A_2}$ and $y_{A_3}$ has two sets of solutions. Therefore, the number of the inverse position problem is $2 \times 2 \times 8 = 32$.

## Numerical simulation for direct and inverse kinematics

### Direct solutions

The dimension parameters of the PM are set to $a=300$, $b=150$, $d=50$, $l_1=30$, $l_2=280$, $l_3=140$, $l_4=180$, $l_5=90$, $l_6=230$ in the unit of mm. Let the length of the connecting links 5 between the parallelograms and the length of the connecting link 8 are set to $l_7=0$ and $l_8=0$, respectively. As an example, if the three input values are $y_{A_1}=350$, $y_{A_2}=-300$, $y_{A_3}=-25$, there are eight real direct kinematic solutions, as listed in Table 1.

### Inverse solutions

In Table 1, the third direct solution is substituted into the Eq. (21), and the 32 inverse kinematic solutions are obtained, as shown in Table 2. It can be seen that the 7$^{th}$ inverse kinematic solutions from Table 2 is consistent with the three input values given when the direct kinematic problem is solved, which means the validation of the procedure of the direct kinematic derivation.

TABLE 1 THE VALUES OF DIRECT SOLUTIONS

| No. | x(mm) | y(mm) | z(mm) |
|---|---|---|---|
| 1 | -123,24178 | 25 | -249,844792 |
| 2 | -29,6299347 | 25 | -4,50975921 |
| 3 | -2,99651958 | 25 | 11,1500571 |
| 4* | 25,2633156 | 25 | 23,019356 |
| 5 | 25,2633156 | 25 | 36,980644 |
| 6 | -2,99651958 | 25 | 48,849943 |
| 7 | -29,6299347 | 25 | 64,5097592 |
| 8 | -123,24178 | 25 | 309,844792 |

**Discussion**

It should be noted that the number of real solutions to the direct and inverse kinematic model is not constant within the workspace. However, they are the maximum number of the direct and inverse kinematics. As a comparison, the number of solutions of a linear Delta robot are only two solutions for the direct kinematic model and height for the inverse kinematic model. If the kinematic model is more complex, joint limits can be easily introduced to avoid singularities. An optimization based on the Jacobian matrices and its isotropic posture can yield a new Orthoglide mechanism where a given assembly mode and a given working mode are fixed [4, 26].

## SINGULARITY ANALYSIS

Singularity analysis can be performed by studying HSOCs separately. From [27], the serial and parallel singularities are investigated. The biglide mechanism depicted in the Fig. 4 is similar to the $HSOC_1$ and the upper part of $HSOC_2$.

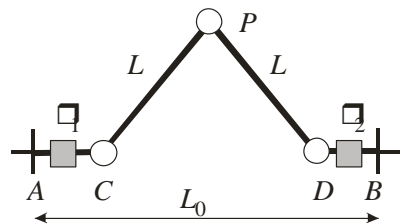

FIGURE 4 THE BIGLIDE MECHANISM

Three types of singularities can be determined from the study of the Jacobian matrix that links joint velocities to Cartesian velocities (i) the serial singularities, (ii) the parallel singularities and (iii) the constraint singularities [28].

When the manipulator is in serial singularities, there is a direction along which no Cartesian velocity can be produced. The serial singularities define the boundaries of the Cartesian workspace [29]. For the mechanism depicted in Fig. 5, the serial singularities occur whenever the axis of the actuated prismatic joint (BD) is orthogonal to the leg (DP). For the mechanism under study, such configurations exist as long as $(B_1C_1)$ or $(B_2C_2)$ or $(B_3C_3)$ or $(D_1D_2)$ or $(D_3E_3)$ are vertical. This property explains the existence of 32 solutions to the inverse kinematic model.

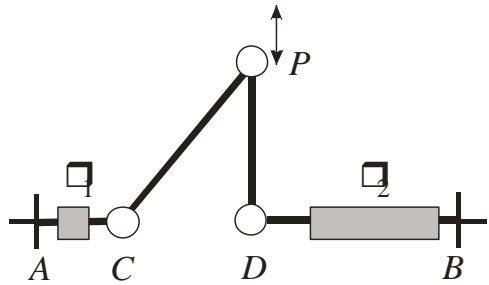

FIGURE 5 THE BIGLIDE MECHANISM, SERIAL SINGULARITY

TABLE 2 THE VALUES OF INVERSE KIENMATICS SOLUTIONS

| No. | $y_{A_1}$ | $y_{A_2}$ | $y_{A_3}$ |
|---|---|---|---|
| 1 | -160 | -300 | -25 |
| 2 | -165,881846 | -305,881846 | -25 |
| 3 | -160 | -300 | -67,5941964 |
| 4 | -165,881846 | -305,881846 | -67,5941964 |
| 5 | -160 | -300 | 75 |
| 6 | -165,881846 | -305,881846 | 75 |
| 7* | 350 | -300 | -25 |
| 8 | 350 | -300 | -67,5941964 |
| 9 | 355,881846 | -305,881846 | -25 |
| 10 | 355,881846 | -305,881846 | -67,5941964 |
| 11 | -160 | -300 | 117,594196 |
| 12 | -165,881846 | -305,881846 | 117,594196 |
| 13 | 350 | -300 | 75 |
| 14 | 355,881846 | -305,881846 | 75 |
| 15 | -160 | 210 | -25 |
| 16 | -160 | 210 | -67,5941964 |
| 17 | -165,881846 | 215,881846 | -25 |
| 18 | -165,881846 | 215,881846 | -67,5941964 |
| 19 | 350 | -300 | 117,594196 |
| 20 | 355,881846 | -305,881846 | 117,594196 |
| 21 | -160 | 210 | 75 |
| 22 | -165,881846 | 215,881846 | 75 |
| 23 | 350 | 210 | -25 |
| 24 | 350 | 210 | -67,5941964 |
| 25 | 355,881846 | 215,881846 | -25 |
| 26 | 355,881846 | 215,881846 | -67,5941964 |
| 27 | -160 | 210 | 117,594196 |
| 28 | -165,881846 | 215,881846 | 117,594196 |
| 29 | 350 | 210 | 75 |
| 30 | 355,881846 | 215,881846 | 75 |
| 31 | 350 | 210 | 117,594196 |
| 32 | 355,881846 | 215,881846 | 117,594196 |

In the parallel singularities, it is possible to move locally the tool center point even though the actuated joints are locked. These singularities are particularly undesired, because the structure cannot resist any external force and the motion of the mechanism is uncontrolled. To avoid any deterioration, it is necessary to eliminate the parallel singularities from the workspace. For the mechanism depicted in Fig. 6, the parallel singularities occur whenever the points *C*, *D*, and *P* are aligned or whenever *C* and *D* coincide.

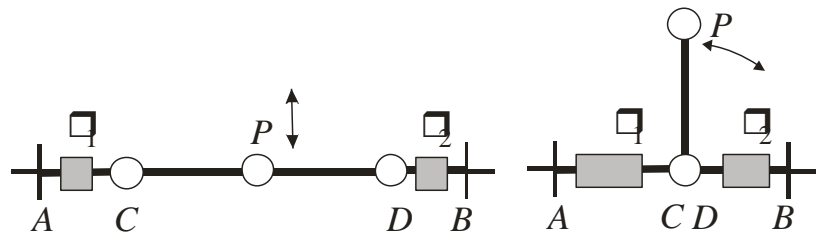

FIGURE 6 THE BIGLIDE MECHANISM, PARALLEL SINGULARITIES

For the mechanism under study, this property is related to $(B_1C_1)$ or $(B_2C_2)$ parallel or aligned, or $(D_1D_2)$ or $(D_3E_3)$ parallel or aligned or $(C_3B_3)$ is parallel to the axis of the prismatic joint $P_3$. There are thus three conditions for parallel singularities, which justifies the existence of eight solutions to the direct kinematic problem. Figures 7 and 8 represent an example of serial and parallel singular configurations, respectively.

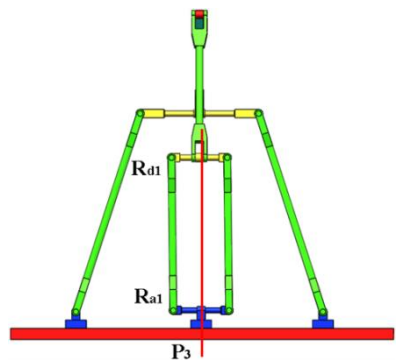

FIGURE 7 EXAMPLE OF SERIAL SINGULARITY CONFIGURATION

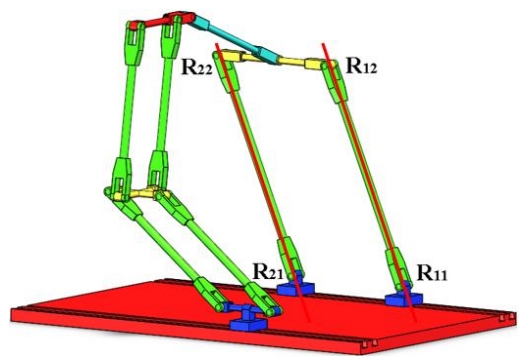

FIGURE 8 EXAMPLE OF PARALLEL SINGULARITY CONFIGURATION

The constraint singularities are related to parallelograms [28, 30] whenever they become anti-parallelogram (or flat parallelogram) and can no longer constrain the two opposite bars to remain parallel. These singularities exist when the revolute joints ($R_{a1}$, $R_{b1}$, $R_{c1}$, $R_{d1}$) or ($R_{a2}$, $R_{b2}$, $R_{c2}$, $R_{d2}$) are aligned.

# CONCLUSIONS

The paper presents a 3-translational parallel mechanism with three main advantages: (1) it is only composed of three actuated prismatic joints and passive revolute joints, which is easy to be manufactured and assembled; (2) its direct and inverse kinematics can be solved analytically, simplifying error analysis, dimensional synthesis, stiffness and dynamics modeling; and (3) it has partial input-output motion decoupling, which is very beneficial to the trajectory planning and motion control of the PM.

According to the kinematic modeling principle proposed by the author based on the single-opened-chains units and topological characteristics method, in the first loop with positive constraint degree, the set one virtual variable $\gamma$ can be directly obtained by the special topological characteristics constraint condition that the output link of the first loop always maintains the horizontal position, where the condition is provided by the second loop with negative constraint degree. Therefore, the entire analytical position solutions are obtained without solving the virtual variable $\gamma$ by the geometric constraint equation in the second loop with negative constraint degree. This is the advantage of the topology of the proposed PM being different from other PMs, and it has analytical direct solutions. The method has clear physical meaning and simple calculation. Thus, the loci of the singular configurations of the PM are identified.

The work of this paper lays the foundation for the stiffness, trajectory planning, motion control, dynamics analysis and prototype design for this new PM.

# ACKNOWLEDGMENTS:

This research is sponsored by the NSFC (Grant No.51475050 and No.51975062) and Jiangsu Key Development Project (No.BE2015043).

# List of Figures

# List of Tables